\documentclass[letterpaper, 10 pt, journal, twoside]{IEEEtran}  

\IEEEoverridecommandlockouts                              

\usepackage{cite}
\usepackage{amssymb,amsfonts}
\usepackage{algorithmic}
\usepackage{graphicx}
\usepackage{subfloat}
\usepackage{textcomp}
\usepackage{xcolor}
\usepackage{algorithm}
\usepackage{verbatim} 
\usepackage{mathrsfs}
\usepackage{subfigure}
\usepackage{empheq}
\usepackage{lipsum}
\usepackage{comment}
\usepackage{mathtools, cuted}
\usepackage{bm}

\usepackage{amsthm}
\usepackage[utf8]{inputenc}
\usepackage[english]{babel}
\usepackage[colorlinks,citecolor=red,urlcolor=blue,bookmarks=false,hypertexnames=true]{hyperref} 
\usepackage[textsize=tiny]{todonotes}
\usepackage[font=footnotesize,skip=4pt]{caption}
\usepackage{etoolbox}
\apptocmd{\thebibliography}{\scriptsize}{}{}

\newlength{\textfloatsepsave} 

\newcommand*{\Revised}{\textcolor{black}}
\usepackage{soul}
\newcommand{\vg}[1]{\bm{#1}}
\renewcommand{\v}[1]{\mathbf{#1}}

\newtheorem*{definition*}{Definition}

\newtheorem{remark}{Remark}
\makeatletter
\newenvironment{formulation}[1][htb]{%
    \renewcommand{\ALG@name}{Formulation}
   \begin{algorithm}[#1]%
  }{\end{algorithm}}
\makeatother

\begin{document}
\title{Momentum-Aware Trajectory Optimization and Control for Agile Quadrupedal Locomotion
}

\author{Ziyi Zhou*,~\IEEEmembership{Student Member,~IEEE}, Bruce Wingo*,~\IEEEmembership{Student Member,~IEEE}, Nathan Boyd,~\IEEEmembership{Student Member,~IEEE}, Seth Hutchinson,~\IEEEmembership{Fellow,~IEEE}, and Ye Zhao,~\IEEEmembership{Member,~IEEE}%
\thanks{This work was supported by NSF IIS-1924978, Georgia Tech Research Institute IRAD Grant.} 
\thanks{*The first two authors equally contributed to this work}
\thanks{Institute for Robotics and Intelligent Machines, Georgia Institute of Technology, Atlanta, GA 30332, USA 
        {\tt\footnotesize \{zhouziyi, bwingo, nboyd31, yezhao, seth\}@gatech.edu}
        }%
\thanks{The authors would like to thank the MIT Biomimetic Robotics Lab and NAVER LABS for providing the Mini Cheetah simulation software and lending the Mini Cheetah for experiments.}
}

\IEEEaftertitletext{\vspace{-1\baselineskip}}
\maketitle

\begin{abstract}
In this letter, we present a versatile hierarchical offline planning algorithm, along with an online control pipeline for agile quadrupedal locomotion.
Our offline planner alternates between optimizing centroidal dynamics for a reduced-order model and whole-body trajectory optimization, with the aim of achieving dynamics consensus.
Our novel momentum-inertia-aware centroidal optimization, which uses an equimomental ellipsoid parameterization, is
able to generate highly acrobatic motions via ``inertia shaping".
Our whole-body optimization approach significantly improves upon the quality of standard DDP-based approaches
by iteratively exploiting feedback from the centroidal level.
\Revised{For online control, we have developed a novel convex model predictive control scheme through a linear transformation of the full centroidal dynamics.}
Our controller can efficiently optimize for both contact forces and joint accelerations in single optimization,
enabling more straightforward tracking for momentum-rich motions compared to existing quadrupedal MPC controllers.
We demonstrate the capability and generality of our trajectory planner on four different dynamic maneuvers.
We then present \Revised{one hardware experiment} on the MIT Mini Cheetah platform to demonstrate the
performance of the entire planning and control pipeline on a twisting jump maneuver.

\end{abstract}

\vspace{-0.1in}

%
\section{Introduction}\label{sec:introduction}

\IEEEPARstart{W}{ith} the recent rapid increase in capabilities of legged robots, the demand for more sophisticated
approaches to trajectory optimization and online control
has motivated a plethora of optimization-based motion planning strategies.
Broadly speaking, these can be divided into methods that solve
a single trajectory optimization (TO) problem,
and those that decompose the overall problem into a set of decoupled subproblems.
Single-optimization methods can be further subdivided into those that use
reduced order models and those that optimize over the full-order dynamics.

\begin{figure}[tb!]
    \centering
    \includegraphics[width=\linewidth]{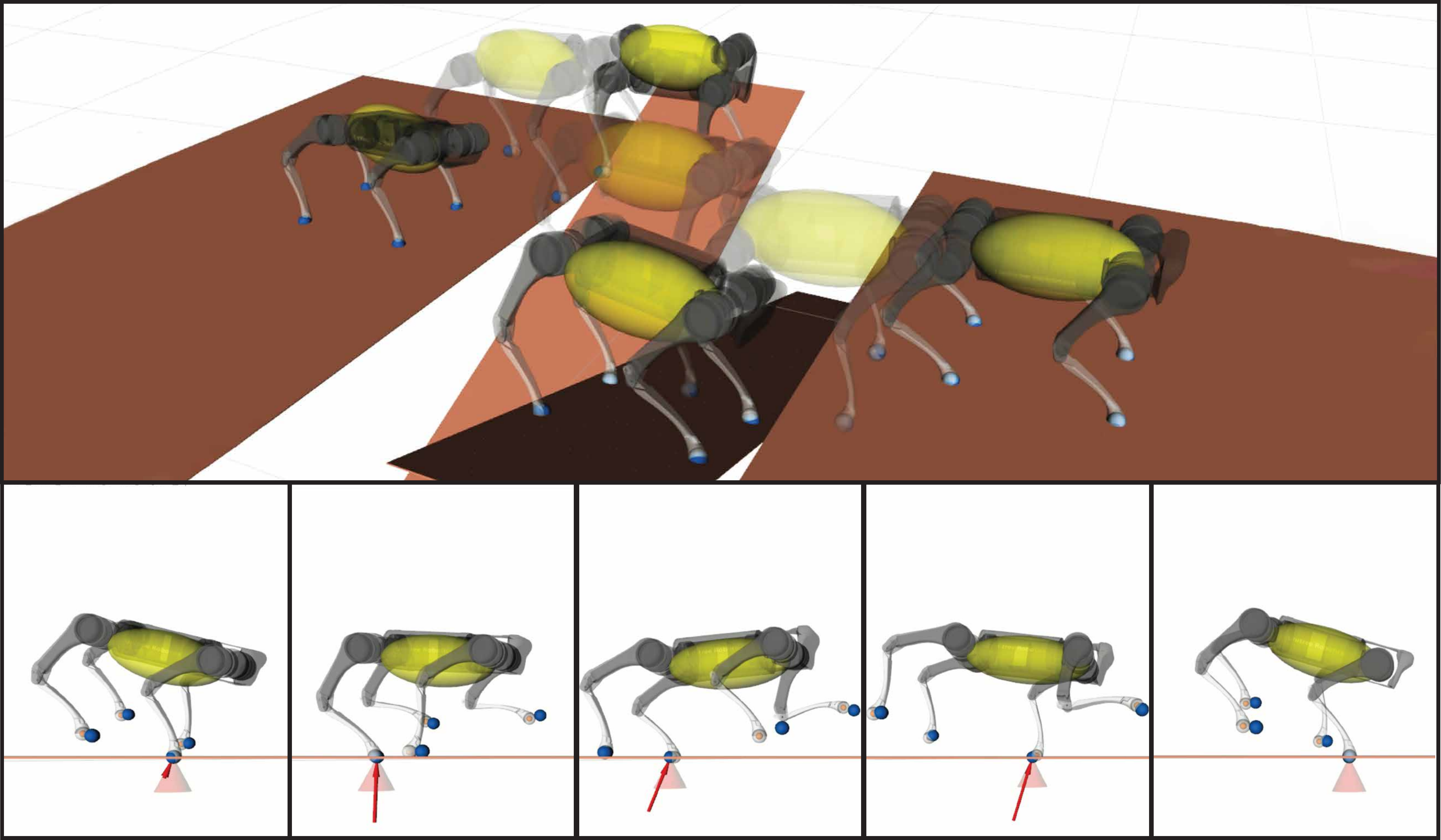}
    \caption{Dynamic maneuvers that require high-fidelity control of angular momentum, such as ``parkour" (top) and cantering (bottom).}
    \label{fig:sim_ex}
    \vspace{-0.25in}
\end{figure}

Reduced-order models, such as the spring loaded inverted pendulum (SLIP) model, single rigid body model (SRBM), centroidal dynamics model, etc., have been widely adopted.
High speed and robust running are demonstrated in \cite{di_carlo_dynamic_2018, kim_highly_2019, bledt_regularized_2020}, using SRBM and user defined heuristics;
however, their framework can only generate cyclic and short-horizon gaits.
To generate more complex motions, Winkler et al. proposed TOWR \cite{winkler_gait_2018}, which is capable of directly optimizing over contact sequences and timings, through its phase-based parameterization.
Incorporating joint information to reason about the centroidal inertia is non-trivial with these models, which limits their ability to generate angular-momentum-rich motions using limb movements \Revised{such as the maneuvers shown in Fig.~\ref{fig:sim_ex}}.
\Revised{TOWR+ \cite{ahn_versatile_2021} and the lumped-leg SRBM \cite{wang_unified_2021} attempted to augment the standard SRBM by approximating the time-varying inertia tensor; however, neither use the full capabilities of the joint motions.}
Earlier works, such as \cite{zordan_control_2014} and \cite{lee_reaction_2007} derived analytical models, known as equimomental ellipsoids, to parameterize centroidal inertia; however, discovering physically meaningful inertia trajectories solely through the ellipsoid parameterization has proven difficult.

It is also possible to use full-order dynamic models in single-optimization approaches.
One option is to use full dynamics with either a hybrid \cite{hereid_3d_2016,mastalli_crocoddyl_2020,li_hybrid_2020} or contact-implicit \cite{posa_direct_2014,neunert_trajectory_2017,tassa_synthesis_2012,mordatch_discovery_2012} formulation. Alternatively, centroidal dynamics are utilized in \cite{dai_whole_2014,farshidian_efficient_2017,sleiman_constraint_2021} with full kinematics through the centroidal momentum matrix (CMM) \cite{orin_centroidal_2013}.
Notably, the differential dynamic programming (DDP)-based indirect approaches, such as FDDP \cite{mastalli_crocoddyl_2020} and constrained SLQ \cite{sleiman_constraint_2021}, have shown impressive results in terms of computational efficiency and constraint handling.
Additional efforts have been made in \cite{tassa_control_2014,xie_differential_2017,howell_altro_2019} to enforce general equality and inequality constraints; however, a both efficient and general-purpose solver, capable of handling nonlinear full-order models, does not yet exist, especially for highly agile acrobatic motions.

\begin{figure*}[t]
    \centering
    \smallskip
    \smallskip
    \smallskip
    \includegraphics[width=\linewidth]{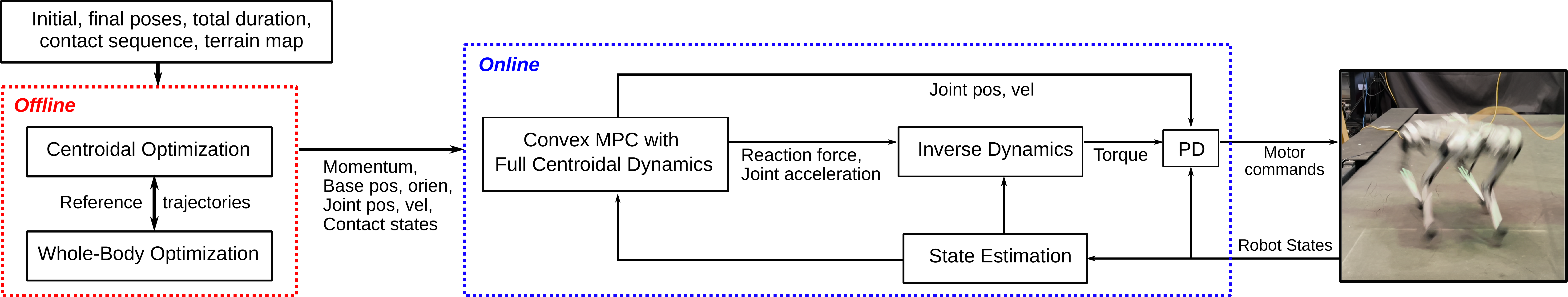}
    \caption{Overall planning and control framework that consists of offline trajectory generation (red) and online execution (blue). The input to offline block is specifically decided by the desired motion. Full-body, centroidal, and contact information are passed from offline generation to online control.}
    \label{fig:system_diagram}
    \vspace{-0.2in}
\end{figure*}

By decomposing trajectory optimization into simpler sub-problems,
it is possible to leverage the benefits of both reduced and full-order models.
One strategy is through a hierarchical optimization.
In \cite{carpentier_multicontact_2018,kudruss_optimal_2015}, a centroidal optimization solves for momentum and contact force trajectories, then inverse kinematics is applied to generate joint motions.
Similarly, Nguyen et al. \cite{nguyen_contact_2021} proposed a pipeline, first solving an SRBM-based centroidal optimization, then using its solution as the tracking objective for a full dynamics-based whole-body optimization.
However, as pointed out by \cite{budhiraja_dynamics_2019}, additional constraints projecting whole-body information to the centroidal level are required to ensure feasible solutions, which are often computationally expensive and difficult to define, especially when involving angular momentum (AM).

An alternative strategy is to iteratively solve the two optimizations in an alternating fashion until the convergence of a common set of dynamics variables is achieved (dynamics consensus) \cite{herzog_structured_2016,ponton_efficient_2021,budhiraja_dynamics_2019,zhou_accelerated_2020}.
To date, existing frameworks compromise in either of the two optimizations: only full kinematics models (instead of dynamics ones) are considered in \cite{herzog_structured_2016,ponton_efficient_2021} during its whole-body optimization, while the centroidal optimizations proposed in \cite{budhiraja_dynamics_2019, zhou_accelerated_2020} did not explicitly optimize over foot placements. To our knowledge, no alternating optimization frameworks currently exist that both optimize over contact and use inertia-aware reduced order models at the centroidal level, while formulating constrained full dynamics DDP at the whole-body level.

In this letter, we introduce a new alternating centroidal and whole-body optimization framework.
Compared to existing alternating frameworks, we improve the centroidal optimization by incorporating the equimomental ellipsoid parameterization of centroidal inertia \cite{lee_reaction_2007} with SRBM,
and by leveraging the inertia feedback from whole-body optimization to enable a wide range of motions previously unachievable with standard centroidal optimization (e.g., ``inertia shaping" \cite{lee_reaction_2007} that aims to reach a goal pose through precise control of centroidal inertia).
 In addition, our whole-body optimization is designed to only incorporate minimal constraints to track nominal trajectories from the centroidal level in its objective; therefore, efficient methods such as DDP can be used.
We ensure solution feasibility, in terms of constraint satisfaction, by handling most task-space constraints at the centroidal level via direct collocation, and improve solution accuracy  at the whole-body level through dynamics consensus.

In order to implement our method on real robots, we have developed a novel discrete-time finite-horizon model predictive tracking controller (MPC).
Many predictive tracking controllers exist in the legged locomotion control literature;
however, to achieve reasonable control frequency, it is common to use reduced order models within the problem formulation \cite{di_carlo_dynamic_2018, bledt_regularized_2020}.
Although efficient and capable, these controllers are primarily designed for stabilizing fast periodic gaits, and often require expertly designed cost heuristics to generate desired motion \cite{bledt_regularized_2020}. These controllers also do not explicitly solve for joint variables, which makes precisely tracking angular-momentum-rich joint motions impossible without introducing additional layers of whole-body kinematics/dynamics optimization \cite{kim_highly_2019, wang_unified_2021}.
Our implementation of MPC aims to address these short comings. Specifically, the SRBM is augmented with centroidal momentum terms, and the joint variables are introduced through the CMM. With some mild assumptions, the proposed model can be linearized, which not only allows for an efficient QP formulation of the MPC similar to \cite{di_carlo_dynamic_2018}, but also grants us direct control over the joint variables for momentum tracking.

Our core contributions are highlighted as follows:
\begin{itemize}
    \item We introduce an alternating centroidal, whole-body optimization scheme targeting a wide range of motions. Inertia shaping tasks can be performed due to our inertia-aware SRBM based centroidal optimization.
We also demonstrate improvement in solution feasibility compared to the standard whole-body optimization, via dynamics consensus.

    \item \Revised{We formulate a novel convex model predictive controller by modifying the original \cite{di_carlo_dynamic_2018} with a linear transformation of the standard full centroidal dynamics.}
    Our MPC is capable of tracking momentum-rich motions by jointly optimizing over contact forces and joint accelerations, while still retaining model simplicity for fast online computation.
     
    \item We show that the proposed \Revised{trajectory optimization pipeline} generates high-fidelity contact and momentum-rich agile maneuvers, and demonstrate the capability of our controller \Revised{through numerical simulations and one experiment performing a 180-degree twisting jump}.
   
\end{itemize}

\section{System Overview}\label{sec:system_overview}


We propose a hierarchical planning and control framework in this letter, as shown in Fig. \ref{fig:system_diagram}. The inputs to our framework consist of the desired motions, specified by the user (e.g. trot, canter, twisting jump, etc.), along with the associated contact sequence, total motion duration, initial and final poses, and the terrain map. The offline trajectory planner solves centroidal and whole-body optimizations in an alternating fashion, until dynamics consensus is achieved. The resulting centroidal momentum, base motion, and joint trajectories are modified and tracked online using the proposed full centroidal MPC, which re-computes foot contact forces and joint accelerations based on feedback from \Revised{an state estimator}. \Revised{The contact forces and joint accelerations are then sent to an inverse dynamics block for computing the feedforward joint torques, which are combined with the feedback term and passed to the PD.} 
\vspace{-0.15in}
\subsection{System Modeling}
Consider the standard floating base model of a legged robot, with an unactuated $6$-DoF base and a set of $n$-DoF fully-actuated limbs. The equations of motion is given by: 
\begin{equation} \label{eq:fbEoM}
    \resizebox{0.6\hsize}{!}{$
    \begin{aligned}
        &\mathbf{M}(\mathbf{q}) \mathbf{\dot{v}} 
        + \mathbf{C}(\mathbf{q, v}) 
        = \underbrace{ \begin{bmatrix} \textbf{0} \\ \textbf{I} \end{bmatrix} }_{\mathbf{B}} \vg \tau
        + \mathbf{J_{\rm c}^{\top}} \mathbf{F_{\rm c}} &&
    \end{aligned}
    $}
\end{equation}
where $\mathbf{q} = \begin{bmatrix} \mathbf{q_{\rm b}}^{\top} & \mathbf{q_{\rm j}}^{\top} \end{bmatrix}^{\top} \in \mathbb{R}^{n_{q}} \vspace{1mm}$, $\mathbf{v} = \begin{bmatrix} \vg{\nu}_{\rm b}^{\top} & \mathbf{\dot{q}_{\rm j}}^{\top} \end{bmatrix}^{\top} \in \mathbb{R}^{n_{v}}$ are the generalized coordinates and velocities partitioned in base and joint variables. The subscripts $\rm b$ and $\rm j$ are base and joint related quantities, respectively. Base coordinates $\mathbf{q_{\rm b}} = \begin{bmatrix} \mathbf{p_{\rm b}}^{\top} & \vg{\theta_{\rm b}}^{\top} \end{bmatrix}^{\top} \in \mathbb{R}^{6}$ are partitioned as base position and orientation. Base velocities $\vg{\nu}_{\rm b} = \begin{bmatrix} \mathbf{\dot{p}_{\rm b}}^{\top} & \vg{\omega}^{\top} \end{bmatrix}^{\top} \in \mathbb{R}^{6} \vspace{1mm}$ are partitioned as base linear and angular velocities in the world frame. $\mathbf{M(q)}$ is the joint space mass matrix. $\mathbf{C(q, v)}$ captures the nonlinear effects. $\vg{\tau} \in \mathbb{R}^{n_{j}}$ denotes joint torques. $\mathbf{J_{\rm c}(q)}$ is the stacked contact Jacobian and $\mathbf{F_{\rm c}}$ is the stacked contact reaction force vector. Without loss of generality, an implicit assumption $\mathbf{v} = \mathbf{\dot{q}}$ is made for the subsequent development of centroidal and whole-body optimization. This model will be referred to as the \textbf{full dynamics} of the robot.


Assuming full control authority over the joint variables $\mathbf{q_{\rm j}}$, Eq. (\ref{eq:fbEoM}) can be converted into Newton-Euler equations of motion about the robot's center-of-mass (CoM) \cite{spong2006robot}:
\begin{equation}\label{eq:cenDyn}
  \resizebox{0.45\hsize}{!}{$ 
    \mathbf{\dot{h}} = \begin{bmatrix} \mathbf{\dot{k}} \\ \mathbf{\dot{l}} \end{bmatrix} 
    = \begin{bmatrix} \sum\limits_{\rm j = 0}^{n_{f}-1}\mathbf{f_{\rm j}} + m \mathbf{g} \\ \sum\limits_{\rm j=0}^{n_{f}-1}(\mathbf{p_{\rm j}} - \mathbf{r}) \times \mathbf{f_{\rm j}} \end{bmatrix}
    $}
\end{equation}
where it is assumed that the robot has $n_f$ feet with point contacts. $\mathbf{h} = \begin{bmatrix} \mathbf{k}^{\top} & \mathbf{l}^{\top}\end{bmatrix}^{\top}$ are the linear and angular centroidal momentum. $\mathbf{p_{\rm{j}}}$ and $\mathbf{r}$ are the individual contact and CoM position in the world frame. $\mathbf{f_{\rm j}}$ are the individual foot contact forces. This reduced-order model will be referred to as the \textbf{centroidal dynamics} throughout this letter. 
\vspace{-0.1in}
\section{Offline Trajectory Generation}\label{sec:trajectoryGeneration}


We present an alternating centroidal, whole-body trajectory optimization scheme in this section. Detailed formulations for individual components are provided in Sec. \ref{subsec:cen_traj_gen} and \ref{subsec:wbd_traj_gen}. The alternating update scheme and the importance of dynamics consensus are discussed in Sec. \ref{subsec:alternating_updates}.

\vspace{-0.1in}
\subsection{Centroidal Trajectory Optimization}\label{subsec:cen_traj_gen}
Formulation \ref{formulation:OCP for centroidal} details the TO problem to be solved at the centroidal level, which utilizes the \textbf{centroidal dynamics} in Eq. (\ref{eq:cenDyn}). We transcribe the optimization problem into a nonlinear program through direct collocation.
Total duration $T$ of the desired motion is equally divided into $N$ knot points. 

The centroidal decision set $\vg \phi_{\rm cen}$ consists of CoM position $\v r$, CoM linear velocity $\dot{\v r}$, angular excursion with Euler angle parameterization $\vg \theta$ and its rate $\dot{\vg \theta}$, centroidal angular momentum $\v l$ and its rate $\dot{\v l}$, $j^{\rm th}$ foot's position $\v p_{\rm j}$ and contact force $\v f_{\rm j}$. 
$\v e$ and $\vg \gamma$ are inertia variables characterizing the principle semi-axes and orientation of the equimomental ellipsoid \cite{lee_reaction_2007} (illustrated in Fig.~\ref{fig:TO_system_diagram}). Ellipsoid orientation is again parameterized using Euler angles. All components of $\vg \phi_{\rm cen}$ belongs to $\mathbb{R}^3$. 
The optimization objective contains a user-defined cost $\mathcal{L}_{\rm cen}(\cdot)$ which specifies various heuristics, such as penalizing the velocity of foot position to discourage violent movements. In addition, a tracking cost $\Psi_{\rm cen}(\vg \phi_{\rm cen},\vg \phi_{\rm wbd}^{\rm ref})$ that minimizes the tracking error to the whole-body reference trajectories (solution to Formulation \ref{formulation:OCP for whole-body} and the variable $\vg \phi_{\rm wbd}^{\rm ref}$ is defined in Formulation \ref{formulation:OCP for centroidal}) are included. Benefits of this tracking cost will be discussed in Sec.~\ref{subsec:alternating_updates}. 
\begin{equation*}
  \resizebox{0.7\hsize}{!}{$ 
\begin{aligned}
     \Psi_{\rm cen} = &\|\v r - \v r^{\rm ref}_{\rm wbd} \|^2_{\mathbf{Q^{\rm cen}_{\rm r}}}+\|m \mathbf{\dot{r}} - \mathbf{k^{\rm ref}_{\rm wbd}} \|^2_{\mathbf{Q^{\rm cen}_{\rm k}}}+\\
     &\|\v l - \v l^{\rm ref}_{\rm wbd} \|^2_{\mathbf{Q^{\rm cen}_{\rm l}}}+\|\v I_{\rm ellip} - \v I_{\rm crb} \|^2_{\mathbf{Q^{\rm cen}_{\rm I}}}+\\&\sum_{j=0}^{n_f-1}(\|\v p_j - \v p_{j,\rm wbd}^{\rm ref} \|^2_{\mathbf{Q^{\rm cen}_{\rm p}}})
\end{aligned}$}
\end{equation*}
Symmetric positive definite weighting matrices for tracking \Revised{the reference} CoM position \Revised{$\v r^{\rm ref}_{\rm wbd}$}, linear momentum (LM) \Revised{$\mathbf{k^{\rm ref}_{\rm wbd}}$}, angular momentum (AM) \Revised{$\v l^{\rm ref}_{\rm wbd}$}, and foot positions \Revised{$\v p_{j,\rm wbd}^{\rm ref}$} are denoted as $\mathbf{Q^{\rm cen}_{\rm r}}$, $\mathbf{Q^{\rm cen}_{\rm k}}$, $\mathbf{Q^{\rm cen}_{\rm l}}$, and $\mathbf{Q^{\rm cen}_{\rm p}}$, respectively. Inertia tracking is achieved through minimizing the $\mathbf{Q^{\rm cen}_{\rm I}}$-weighted Frobenius norm of the difference between the inertia tensors of the equimomental ellipsoid $\mathbf{I_{\rm ellip}}$ and that of the reference composite rigid body (CRB) $\mathbf{I_{\rm crb}}$, both expressed in the world frame. Constraints in centroidal optimization include:
 
$\mathbf{Dynamic \ constraint}$:
The discrete-time non-linear \textbf{centroidal dynamics} are used. Eq.~(\ref{eq:dynamics}) explicitly includes the relationship between centroidal angular momentum, ellipsoid inertia tensor, and CoM angular velocity. The inertia tensor $\v I_{\rm ellip}$ in the world frame is constructed from the rotational transformation of the inertia tensor in the base frame, $\mathbf{I_{\rm ellip}} = \mathbf{R}(\vg \gamma) _{\mathcal{B}}\v I_{\rm ellip}(\mathbf{e}) \mathbf{R}(\vg \gamma)^{\top}$ with $\mathbf{R}(\vg \gamma)$ being the rotation transformation between the ellipsoid base frame and world frame. Inertia tensor in the base frame is defined as $_{\mathcal{B}}\v I_{\rm ellip} = diag([\mathbf{I}^{\rm xx}, \mathbf{I}^{\rm yy}, \mathbf{I}^{\rm zz}]^T)$  with $\mathbf{I}^{xx} = \frac{1}{5}m(\v {e^{\rm y}}^2 + \v {e^{\rm z}}^2)$, $\mathbf{I}^{yy} = \frac{1}{5}m(\v {e^{\rm x}}^2 + \v {e^{\rm z}}^2)$, and $\mathbf{I}^{zz} = \frac{1}{5}m(\v {e^{\rm x}}^2 + \v {e^{\rm y}}^2)$. 

$\mathbf{Ellipsoid \ mass \ constraint}$: To ensure the equimomental ellipsoid has the same mass as the robot, ellipsoid semi-axes, $\mathbf{e}$, must be constrained. Ellipsoid mass $m = \frac{4}{3} \pi \mathbf{e^{\rm x}} \mathbf{e^{\rm y}} \mathbf{e^{\rm z}} \rho$ with $\rho$ being the average mass density. $\rho$ can be obtained from the reference CRB inertia tensor $\mathbf{I_{\rm crb}}$ for each time step \cite{orin_centroidal_2013}. When the inertia reference is unavailable, $\rho$ is kept constant using the initial robot configuration.   

$\mathbf{Integration \ constraint}$:
By adopting trapezoidal integration scheme, the quantities $\v r$, $\dot{\v r}$, $\vg \theta$ and $\v l$ are approximated as piecewise linear functions as shown in Eqs.~(\ref{eq:integration_r}) - (\ref{eq:integration_L}). The final knot points are set to be free but subject to final conditions if needed. $\Delta t$ denotes one time step.

$\mathbf{Frictional \ constraint}$:
Friction cone constraint and unilateral constraint on the contact forces are both modeled in Eqs.~(\ref{eq:force_non_zero}) - (\ref{eq:force_zero}). Specifically, friction cones are approximated as friction pyramids $\mathcal{F}$ to remove nonlinearity. The contact normal forces are non-negative, and always equal to zero during the swing phase.

$\mathbf{Terrain \ constraint}$:
For each foot during the stance phase, Eq.~(\ref{eq:terrain_1}) defines the no-slip condition. While Eq.~(\ref{eq:terrain_2}) guarantees that foot placement remains on the terrain surface, as specified by the height map $h_{\rm terrain}$ (see \cite{winkler_gait_2018} for details).

$\mathbf{Kinematic \ constraint}$:
Foot position range-of-motion constraints, Eq.~(\ref{eq:rom}), are defined, to heuristically bound each foot's movement within a fixed sized box $\mathcal{R}$.

\setlength{\textfloatsep}{5pt}
\begin{subfigures}
\begin{figure}
    \vspace{-0.12in}
\end{figure}
\begin{formulation}[t]
\caption{Centroidal trajectory optimization}\label{formulation:OCP for centroidal}
\begin{subequations}
\resizebox{1.0\linewidth}{!}{
  \begin{minipage}{\linewidth}
    \begin{align}
        \nonumber &\hspace{2.1cm} \underset{\vg \phi_{\rm cen}}{\min} \ \mathcal{L}_{\rm cen}(\vg \phi_{\rm cen}) + \Psi_{\rm cen}(\vg \phi_{\rm cen},\vg \phi_{\rm wbd}^{\rm ref})
        \\\nonumber
        &(\textit{\rm Variables}) \hspace{0.5cm} \resizebox{0.65\linewidth}{!}{$\vg \phi_{\rm cen}[i] = [\v r[i]^{\top}, \dot{\v r}[i]^{\top}, \vg \theta[i]^{\top}, \dot{\vg \theta}[i]^{\top},\v l[i]^{\top},$} \\\nonumber
        &\hspace{3.5cm}  \resizebox{0.5\linewidth}{!}{$\dot{\v l}[i]^{\top}, \v p_j[i]^{\top}, \v f_j[i]^{\top},\underbrace{\v e[i]^{\top}, \vg \gamma[i]^{\top}}_{\text{inertia vectors}}
        ]^{\top}$}\\\nonumber & \hspace{2.1cm} \resizebox{0.6\linewidth}{!}{$\vg \phi_{\rm wbd}^{\rm ref}[i]=[\v r^{\rm ref}_{\rm wbd}[i]^{\top},\mathbf{\dot{r}^{\rm ref}_{\rm wbd}}[i]^{\top},\v l^{\rm ref}_{\rm wbd}[i]^{\top},$}\\
        \nonumber & \hspace{3.7cm} \resizebox{0.3\linewidth}{!}{$\v p^{\rm ref}_{j, \rm wbd}[i]^{\top}, \v I_{\rm crb}[i]^{\top}]^{\top}$}\\\nonumber
        &\hspace{2.1cm} \resizebox{0.5\linewidth}{!}{$\forall i = 0, \ldots, N-1, j=0,\ldots,n_f-1$}\\\label{eq:dynamics}
        &(\textit{\rm Dynamics}) \ \text{s.t.}  \resizebox{0.55\linewidth}{!}{$\begin{bmatrix} m \ddot{\v r}[i]\\\
        \dot{\v l}[i]\\\
        \v l[i]
        \end{bmatrix}= \begin{bmatrix}
        \sum\nolimits_{j} \v f_j[i] + m \v g \\
	    \sum\nolimits_j (\v{p}_j[i] - \v r[i]) \times \v f_j[i] \\
	    \v I_{\rm ellip}(\v e[i], \vg \gamma[i])\vg \omega[i]
	    \end{bmatrix}$} \\\label{eq:ellipsoidMass}
	    &(\textit{\rm Ellipsoid}) \hspace{0.52cm} \resizebox{0.45\linewidth}{!}{$m = \frac{4}{3} \pi \mathbf{e^{\rm x}}[i] \mathbf{e^{\rm y}}[i] \mathbf{e^{\rm z}}[i] \rho[i] $}\\\label{eq:integration_r}
        &(\textit{\rm Integration}) \hspace{0.3cm} \resizebox{0.55\linewidth}{!}{$\v r[i+1] - \v r[i] = \frac{\Delta t}{2}(\dot{\v r}[i+1]+\dot{\v r}[i])$}\\\label{eq:integration_r_dot}
        &\hspace{2.1cm} \resizebox{0.55\linewidth}{!}{$\dot{\v r}[i+1] - \dot{\v r}[i] = \frac{\Delta t}{2}(\ddot{\v r}[i+1]+\ddot{\v r}[i])$}\\\label{eq:integration_theta}
        &\hspace{2.1cm} \resizebox{0.55\linewidth}{!}{$\vg \theta[i+1] - \vg \theta[i] = \frac{\Delta t}{2}(\dot{\vg \theta}[i+1]+\dot{\vg \theta}[i])$}\\\label{eq:integration_L}
        &\hspace{2.1cm} \resizebox{0.55\linewidth}{!}{$\v l[i+1] - \v l[i] = \frac{\Delta t}{2}(\dot{\v l}[i+1]+\dot{\v l}[i])$}\\\label{eq:force_non_zero}
        &(\textit{\rm Frictional}) 
    \hspace{0.45cm} \resizebox{0.48\linewidth}{!}{$\v f_j[i] \cdot \v n(\v p_j^{xy}[i]) \ge 0, \ \forall i \in \mathcal{C}_j$}\\\label{eq:force_friction_cone}
    &\hspace{2.1cm} \resizebox{0.5\linewidth}{!}{$\v f_j[i] \in \mathcal{F}(\mu, \v n, \v p_j^{xy}[i]), \ \forall i \in \mathcal{C}_j$}\\\label{eq:force_zero}
     &\hspace{2.1cm} \resizebox{0.3\linewidth}{!}{$\v f_j[i] = 0, \ \forall i \notin \mathcal{C}_j$}\\\label{eq:terrain_1}
        &(\textit{\rm Terrain}) 
    \hspace{0.82cm} \resizebox{0.5\linewidth}{!}{$\v p_j^{xy}[i+1] - \v p_j^{xy}[i]=0, \ \forall i \in \mathcal{C}_j$}\\\label{eq:terrain_2}
    &\hspace{2.12cm} \resizebox{0.5\linewidth}{!}{$\v p_j^z[i]=h_{\rm terrain}(\v p_j^{xy}[i]), \ \forall i \in \mathcal{C}_j$}\\\label{eq:rom}
        &(\textit{\rm Kinematics}) \hspace{0.22cm} \resizebox{0.3\linewidth}{!}{$\v p_j[i] \in \mathcal{R}_j(\v r[i],\vg \theta[i])$}
    \end{align}
     \end{minipage}
     }
\end{subequations}
\end{formulation}
\end{subfigures}

\vspace{-0.1in}
\subsection{Whole-Body Trajectory Optimization}\label{subsec:wbd_traj_gen}
For legged systems, leveraging full-body dynamics, such as the angular momentum of limbs, is critical for generating dynamic motions. The whole-body optimization is formulated in a hybrid dynamics fashion. The objective is to track quantities computed from centroidal optimization. Formulation \ref{formulation:OCP for whole-body} details this whole-body optimization.

The decision set $\vg \phi_{\rm wbd}$ consists of the generalized position $\v q$ and velocity $\dot{\v q}$, joint torques $\vg \tau$ and contact forces $\mathbf{F_{\rm c}}$. Similar to the centroidal optimization, the objective includes a user-defined heuristic cost $\mathcal{L}_{\rm wbd}$, such as torque minimization, and a tracking cost $\Psi_{\rm wbd}(\vg \phi_{\rm wbd},\vg \phi_{\rm cen}^{\rm ref})$ to minimize the deviation from the centroidal reference trajectories (solution to Formulation \ref{formulation:OCP for centroidal}). 
\begin{equation*}
  \resizebox{0.75\hsize}{!}{$ 
    \begin{aligned}
      \Psi_{\rm wbd}=& \|\mathcal{X}_r(\v q) - \v r^{\rm ref}_{\rm cen} \|^2_{\mathbf{Q^{\rm wbd}_{\rm r}}}+\|\v A(\v q)\dot{\v q} - \v h^{\rm ref}_{\rm cen}\|^2_{\mathbf{Q^{\rm wbd}_{\rm h}}}+\\&\sum_{j=0}^{n_f}(\|\mathcal{X}_j(\v q) - \v p^{\rm ref}_{j,\rm cen}\|^2_{\mathbf{Q^{\rm wbd}_{\rm p}}})
    \end{aligned}$}
\end{equation*}
where the quadratic weighting matrices for tracking \Revised{the reference} CoM position \Revised{$\v r^{\rm ref}_{\rm cen}$}, centroidal momentum \Revised{$\v h^{\rm ref}_{\rm cen}$}, and foot positions \Revised{$\v p^{\rm ref}_{j,\rm cen}$} are denoted as $\mathbf{Q^{\rm wbd}_{\rm r}}$, $\mathbf{Q^{\rm wbd}_{\rm h}}$, and $\mathbf{Q^{\rm wbd}_{\rm p}}$, respectively. $\mathcal{X}_r(\cdot)$ and $\mathcal{X}_j(\cdot)$ map the generalized position to CoM position and the $j \rm th$ end-effector (EE)'s position. The centroidal momentum matrix, $\mathbf{A(q)} \in \mathbb{R}^{6 \times (6 + n_j)}$, referring to Eq. (\ref{eq:cmmToGeneralizedVel}), provides a linear relationship between the centoridal momentum $\mathbf{h}$ and generalized velocities $\mathbf{\dot{q}}$ \cite{orin_centroidal_2013}. \Revised{In Eq.~(\ref{eq:hybrid_dynamics})}, the hybrid dynamics constraint includes \Revised{either the continuous} \textbf{full dynamics} \Revised{(first row)}, \Revised{or} the impact \Revised{dynamics} \Revised{(second row)}, where $\vg \Lambda$ stands for the contact impulse, and $\dot{\v q}^-$ and $\dot{\v q}^+$ are instantaneous joint velocities before and after the impact. Following (\ref{eq:hybrid_dynamics}), a contact constraint (\ref{eq:contact}) is added for the stance \Revised{or impact foot assuming a rigid contact, while a zero force is applied to the non-contact foot.} Joint limit, torque limit, and friction cone are also modeled as shown in Eqs. (\ref{eq:joint_torque_limits}) - (\ref{eq:wbd_friction_cone}).

\begin{subfigures}
\begin{figure}
    \vspace{-0.12in}
\end{figure}
\begin{formulation}[t]
\caption{Whole-body trajectory optimization}
\label{formulation:OCP for whole-body}
\begin{subequations}
\resizebox{1.0\linewidth}{!}{
  \begin{minipage}{\linewidth}
    \begin{align}
        \nonumber &\hspace{1.95cm} \underset{\vg \phi_{\rm wbd}}{\min} \ \mathcal{L}_{\rm wbd}(\vg \phi_{\rm wbd}) + \Psi_{\rm wbd}(\vg \phi_{\rm wbd},\vg \phi_{\rm cen}^{\rm ref})\\\nonumber
        &(\textit{\rm Variables}) \quad \vg \phi_{\rm wbd}[i] = [\v q[i]^{\top}, \dot{\v q}[i]^{\top}, \vg \tau[i]^{\top}, \mathbf{F_{\rm c}}[i]^{\top}]^{\top}\\\nonumber & \hspace{1.95cm} \vg \phi_{\rm cen}^{\rm ref}[i]=[\v r^{\rm ref}_{\rm cen}[i]^{\top},\v h^{\rm ref}_{\rm cen}[i]^{\top},\v p^{\rm ref}_{j,\rm cen}[i]^{\top}]^{\top}\\\nonumber
        &\hspace{1.95cm} \forall i = 0, 1, \ldots, N-1, j=0,\ldots,n_f-1\\\label{eq:hybrid_dynamics}
        &(\textit{\rm Dynamics}) \quad         \text{s.t.} \resizebox{0.6\hsize}{!}{$\begin{cases} 
        \v M(\v q) \v {\ddot q} + \v C(\v q, \v {\dot q})  = \v B \vg \tau + \mathbf{J_{\rm c}^{\top}} \mathbf{F_{\rm c}}\\
        \v M(\v q)\v{\dot{q}}^+ - \v M(\v q)\v{\dot{q}}^- = \mathbf{J_{\rm c}^{\top}} \vg \Lambda
        \end{cases}$}\\\label{eq:contact}
        &(\textit{\rm Contact}) \hspace{1.1cm} \resizebox{0.3\hsize}{!}{$\begin{cases}
        \v J_c \v{\ddot{q}}+\mathbf{\dot{J}_{\rm c}}\v{\dot{q}} = 0\\
        \textcolor{black}{\v J_c \v{\dot{q}}^+ = 0}
        \end{cases}$}\\\label{eq:joint_torque_limits}
        &(\textit{\rm Limits}) \hspace{1.55cm} \resizebox{0.2\hsize}{!}{$\v q \in \mathcal{J}, \vg \tau \in \mathcal{T}$}
        \\\label{eq:wbd_friction_cone}
        &(\textit{\rm Friction cone}) \hspace{0.55cm} \resizebox{0.1\hsize}{!}{$\vg \lambda_c \in \mathcal{K}$}
    \end{align}
    \end{minipage}
}
\end{subequations}
\end{formulation}
\end{subfigures}


Considering the computational efficiency, we then choose the DDP-based method implemented in \cite{mastalli_crocoddyl_2020} to solve the whole-body optimization. Briefly, \Revised{depending on the continuous (impact) phase,} the dynamics and contact constraints are coupled to derive the joint acceleration \Revised{$\v{\ddot{q}}$ (joint velocity after impact $\v{\dot{q}}^{+}$)} and contact forces \Revised{$\v F_c$ (contact impulse $\vg \Lambda$)} simultaneously, so that the forward pass for DDP can be performed. The joint limit, torque limit and friction cone constraints are encoded as quadratic barrier functions inside the objective function. 

\vspace{-0.1in}
\subsection{Alternating CEN-WBD Optimization}\label{subsec:alternating_updates}
\begin{figure}[htb!]
    \centering
    \includegraphics[width=3.0in]{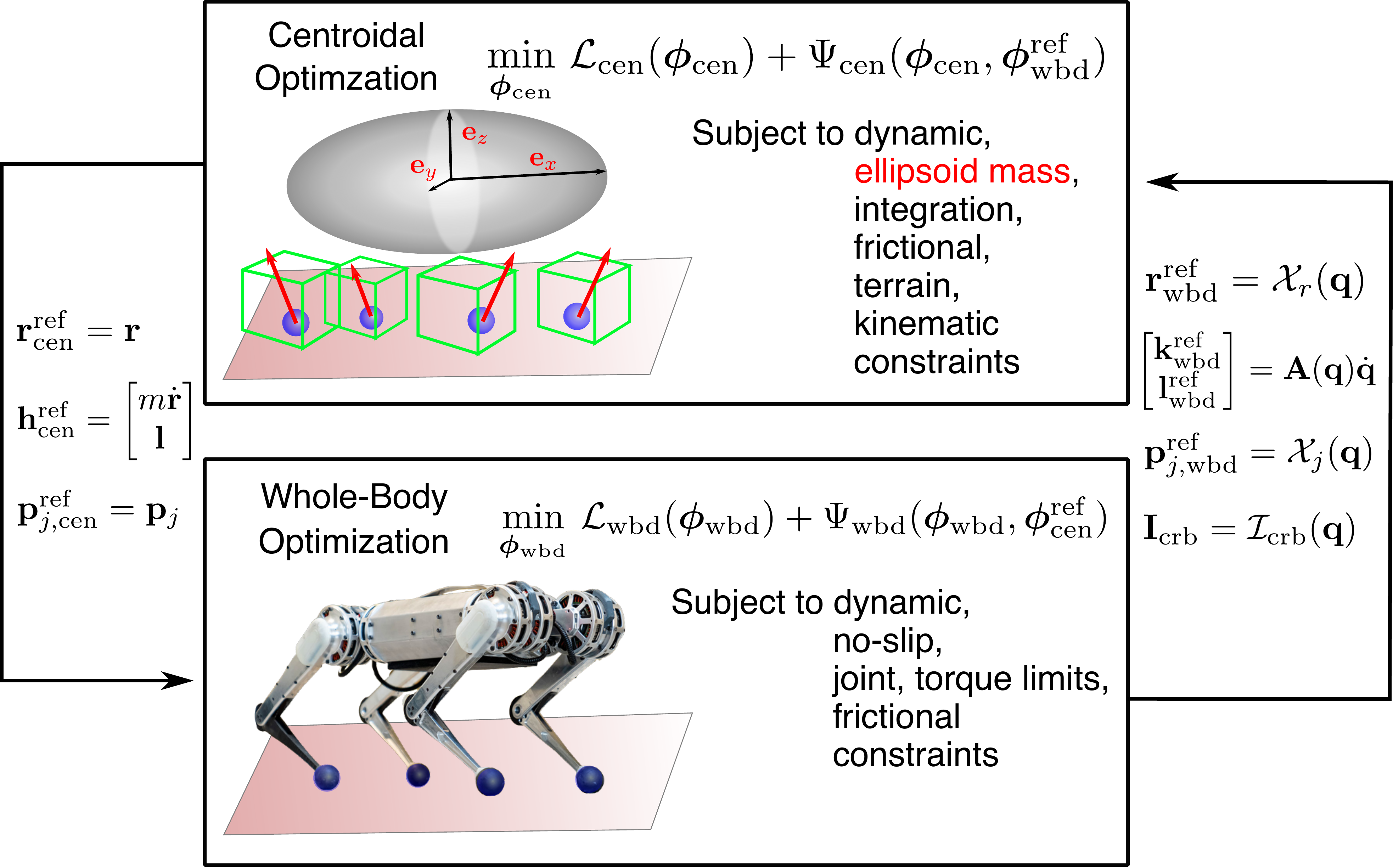}
    \caption{The block diagram that demonstrates the proposed alternating centroidal, whole-body optimization. }
    \label{fig:TO_system_diagram}
    \vspace{-0.2in}
\end{figure}

We perform an alternating update scheme based on the aforementioned centroidal optimization (CEN) and whole-body optimization (WBD) problems. Fig.~\ref{fig:TO_system_diagram} demonstrates the proposed updating sequence, where the CEN optimization is first solved and then followed by the WBD optimization. \Revised{Note that no hand-crafted reference trajectory is required and a constant MoI is employed for the first CEN optimization.}
The \Revised{WBD or CEN} reference trajectories are updated accordingly after each CEN or WBD optimization solve. $\mathcal{I}_{\rm crb}$ maps the generalized coordinates to the CRB inertia tensor $\v I_{\rm crb}$. Similar to \cite{herzog_structured_2016,ponton_efficient_2021,budhiraja_dynamics_2019,zhou_accelerated_2020}, this process, referred to as one alternating iteration, is iterated until the dynamics consensus between CEN and WBD is achieved. Moreover, our proposed consensus quantities include not only CoM position, centroidal momentum, and EE locations, but also the equimomental inertia tensors ($\v I_{\rm ellip}$ in CEN and $\v I_{\rm crb}$ in WBD).  This is achieved by enabling the variables, constraints, and costs that involve ellipsoid inertia vectors $\v e$ and $\v v$ in CEN, which allows generating robot behaviors that cannot be captured by only using centroidal dynamics such as zero-gravity reorientation (see Remark \ref{remark_1}). For the other motions unnecessary to parameterize centroidal inertia, we disable the ellipsoid related terms in CEN and directly use $\v I_{\rm crb}$ from WBD. This hybrid formulation reduces the complexity of directly solving Formulation \ref{formulation:OCP for centroidal} and saves the solve time. More results will be shown in Sec.~\ref{sec:experiments}.

\setlength{\textfloatsep}{\textfloatsepsave}

\begin{remark}
\label{remark_1}
We introduce two sets of orientation representations in CEN: angular excursion $\vg \theta$ and equimomental ellipsoid orientation $\v v$. The former one can be seen as the angular analog of CoM and only changed under external torques \cite{zordan_control_2014}, while the latter one can be affected given any limb movements even without external forces \cite{lee_reaction_2007}.  
\end{remark}

\section{Full Centroidal Convex MPC}
Starting with the derivation of the linearized full centroidal dynamics, this section presents a QP-based MPC problem. We then discuss the benefits of our controller and provides the comparison with the original convex-MPC \cite{di_carlo_dynamic_2018} in Sec. \ref{sec:comparisonOriginalMPC}. 
\vspace{-0.1in}
\subsection{Full Centroidal Dynamics}

Standard full centroidal dynamics as shown in Eq. (\ref{eq:cenDyn}) has been widely used in the legged systems literature \cite{dai_whole_2014, farshidian_efficient_2017}. We propose a linearized variant of this model, and demonstrate its benefits within the context of MPC. To make this linear model unambiguous, the transformation \Revised{$\mathbf{T_{\rm b}}$} between the time derivative of generalized base coordinates and the generalized base velocities is explicitly stated here: 
\begin{equation}\label{eq:quatRateAnglVel}
  \resizebox{0.35\hsize}{!}{$ 
    \mathbf{\dot{q}_{\rm b}} = \underbrace{\begin{bmatrix} \textbf{I} & \textbf{0}_{3 \times 3} \\ \textbf{0}_{4 \times 3} & \mathbf{W}(\vg{\theta}_{\rm b}) \end{bmatrix}}_{\mathbf{T_{\rm b}}(\mathbf{q})} \vg{\nu}_{\rm b}
    $}
\end{equation}
where $\mathbf{W(\vg \theta_{\rm b})}$ represents the velocity transform for different choices of orientation parameterization, see \cite{robotCheatSheet}.

Using the CMM, $\mathbf{A(q)} \in \mathbb{R}^{6 \times (6 + n_j)}$, to relate centroidal momentum to generalized velocities:
\begin{equation}\label{eq:cmmToGeneralizedVel}
  \resizebox{0.35\hsize}{!}{$ 
    \mathbf{h} = \underbrace{\begin{bmatrix} \mathbf{A_{\rm b}}(\mathbf{q}) & \mathbf{A_{\rm j}}(\mathbf{q}) \end{bmatrix}}_{\mathbf{A}(\mathbf{q})} \begin{bmatrix} \vg{\nu}_{\rm b} \\ \mathbf{\dot{q}_{\rm j}} \end{bmatrix}
$}
\end{equation}
Rearranging Eq. (\ref{eq:cmmToGeneralizedVel}) for $\vg{\nu_{\rm b}}$ and using Eq. (\ref{eq:quatRateAnglVel}), one has
\begin{equation}\label{eq:baseCoordRate}
  \resizebox{0.7\hsize}{!}{$
    \mathbf{\dot{q}_{\rm b}} = [~\underbrace{\mathbf{T_{\rm b}}(\mathbf{q}) \mathbf{A_{\rm b}^{-1}}(\mathbf{q})}_{\mathbf{\widetilde{A_{\rm h}}}(\mathbf{q})}~~~\underbrace{-\mathbf{T_{\rm b}}(\mathbf{q}) \mathbf{A_{\rm b}^{-1}}(\mathbf{q}) \mathbf{A_{\rm j}}(\mathbf{q})}_{\mathbf{\widetilde{A_{\rm j}}}(\mathbf{q})}~] \begin{bmatrix} \mathbf{h} \\ \mathbf{\dot{q}_{\rm j}} \end{bmatrix}
  $}
\end{equation}
where $\mathbf{\widetilde{A_{\rm h}}}(\mathbf{q})$ and $\mathbf{\widetilde{A_{\rm j}}}(\mathbf{q})$ will be referred to as transformed CMMs. Defining an augmented state $\mathbf{x} = \begin{bmatrix} \mathbf{h}^{\top},\mathbf{q_{\rm b}}^{\top},\mathbf{q_{\rm j}}^{\top},\mathbf{\dot{q}_{\rm j}}^{\top},\mathbf{g}^{\top} \end{bmatrix}^{\top} \vspace{1mm}$, and combining Eqs. (\ref{eq:cenDyn}) and (\ref{eq:baseCoordRate}), the augmented state dynamics can be written as: 
\begin{equation}\label{eq:fullCenDyn}
  \resizebox{1\hsize}{!}{$ 
  \begin{aligned}
    \begin{bmatrix} \mathbf{\dot{h}} \\ \mathbf{\dot{q}_{\rm b}} \\ \mathbf{\dot{q}_{\rm j}} \\ \mathbf{\ddot{q}_{\rm j}} \\ \mathbf{\dot{g}} \end{bmatrix} 
    &= \underbrace{\begin{bmatrix} \textbf{0} & \textbf{0} & \textbf{0} & \textbf{0} & m \begin{bmatrix} \textbf{I} \\ \textbf{0} \end{bmatrix} \\
    \mathbf{\widetilde{A_{\rm h}}}(\mathbf{q}) & \textbf{0} & \textbf{0} & \mathbf{\widetilde{A_{\rm j}}}(\mathbf{q}) & \textbf{0} \\ \textbf{0} & \textbf{0} & \textbf{0} & \textbf{I} & \textbf{0} \\ \textbf{0} & \textbf{0} & \textbf{0} & \textbf{0} & \textbf{0} \\ \textbf{0} & \textbf{0} & \textbf{0} & \textbf{0} & \textbf{0} \end{bmatrix}}_{\mathbf{H}(\mathbf{q})} 
      \begin{bmatrix} \mathbf{h} \\ \mathbf{q_{\rm b}} \\ \mathbf{q_{\rm j}} \\ \mathbf{\dot{q}_{\rm j}} \\ \mathbf{g} \end{bmatrix} 
    +  \underbrace{\begin{bmatrix} \textbf{I} & \cdots & \textbf{I} & \textbf{0} \\ \widehat{(\mathbf{p_{\rm 0}}-\mathbf{r})} & \cdots & \widehat{(\mathbf{p_{\rm n_f - 1}}-\mathbf{r})} & \textbf{0} \\ \textbf{0} & \cdots & \textbf{0} & \textbf{0} \\ 
    \textbf{0} & \cdots & \textbf{0} & \textbf{0} \\
    \textbf{0} & \cdots & \textbf{0} & \textbf{I} \\
    \textbf{0} & \cdots & \textbf{0} & \textbf{0}
    \end{bmatrix}}_{\mathbf{G}(\mathbf{p}, \mathbf{r})}
    \underbrace{
    \begin{bmatrix}
        \mathbf{f_{\rm 0}} \\ 
        \vdots \\
        \mathbf{f_{\rm n_f - 1}} \\
        \mathbf{a}
    \end{bmatrix}
    }_{\mathbf{u}}
  \end{aligned}
  $}
\end{equation}
We refer to this system as the \textbf{full centroidal dynamics}. The $\widehat{\cdot}$ operator transforms the cross-product operation into a matrix multiplication. Note that, $\mathbf{H}$ only depends on the joint configuration $\mathbf{q}$, and $\mathbf{G}$ only depends on contact and CoM locations, $\mathbf{p}$ and $\mathbf{r}$. \Revised{Each control vector $\mathbf{u}$ consists of the foot contact forces $\mathbf{f}_{j}$ and the joint accelerations $\mathbf{a} \in \mathbb{R}^{n_j}$.} \Revised{Similar to \cite{di_carlo_dynamic_2018}, we assume that} the robot follows the desired reference trajectory generated from our offline optimization, then foot placements, CoM location, and transformed CMMs from offline trajectory can be directly substituted into Eq. (\ref{eq:fullCenDyn}).
\begin{equation} \label{eq:LTVfullCenDyn}
  \resizebox{0.45\hsize}{!}{$ 
    \mathbf{\dot{x}} = \mathbf{H(q^{\rm ref})} \mathbf{x} + \mathbf{G(p^{\rm ref}, r^{\rm ref})} \mathbf{u}
$}
\end{equation}
which simplifies to a linear time-varying (LTV) system in the augmented state $\mathbf{x}$. 


\vspace{-0.1in}
\subsection{Full Centroidal Convex MPC} \label{sec:fullCenConvexMPC}


The discretized version of continuous LTV dynamics Eq. (\ref{eq:LTVfullCenDyn}) is given as:
\begin{equation} \label{eq:DiscreteLTVfullCenDyn}
  \resizebox{0.4\hsize}{!}{$
    \mathbf{x}\scalebox{0.8}{$[i+1]$} = \mathbf{\widetilde{H}}\scalebox{0.8}{$[i]$} \mathbf{x}\scalebox{0.8}{$[i]$} + \mathbf{\widetilde{G}}\scalebox{0.8}{$[i]$} \mathbf{u}\scalebox{0.8}{$[i]$}
  $}
\end{equation}
where $\mathbf{\widetilde{H}}$ and $\mathbf{\widetilde{G}}$ are matrix exponential of $\mathbf{H}$ and $\mathbf{G}$ from Eq. (\ref{eq:LTVfullCenDyn}). Similar to the original convex-MPC \cite{di_carlo_dynamic_2018}, to reduce the size of the proposed MPC problem, state variables can be eliminated from the decision set by adopting a condensed dynamics formulation \cite{jerez_denseQP_2011}. At time step $k$, assuming a MPC prediction horizon of $n \in \mathbb{N}^{+}$ and aggregating Eq. (\ref{eq:DiscreteLTVfullCenDyn}) from $i = k$ to $i = k+n-1$ produces:
\begin{equation}
  \resizebox{0.3\hsize}{!}{$
    \Vec{\mathbf{x}} = \mathbf{\widetilde{H^{\rm \mathcal{A}}}} \mathbf{x}[k] + \mathbf{\widetilde{G^{\rm \mathcal{A}}}} \Vec{\mathbf{u}}
  $}
\end{equation}
with aggregated states, \scalebox{0.8}{}$\Vec{\mathbf{x}} = \scalebox{1.3}{[} \mathbf{x}[k+1]^{\top} ~~ \mathbf{x}[k+2]^{\top} ~ \cdots ~ \mathbf{x}[k+n]^{\top} \scalebox{1.3}{]}^{\top}$. Weighting matrix $\mathbf{\widetilde{H^{\mathcal{A}}}}$ is given as
\begin{equation*}
    \resizebox{0.7\hsize}{!}{$
    \mathbf{\widetilde{H^{\mathcal{A}}}} = \begin{bmatrix} \prod\limits_{i = k}^{k}\mathbf{\widetilde{H}}[i]^{\top} & \prod\limits_{i = k}^{k+1}\mathbf{\widetilde{H}}[i]^{\top} & \cdots & \prod\limits_{i = k}^{k+n-1}\mathbf{\widetilde{H}}[i]^{\top} \end{bmatrix}^{\top}
    $}
\end{equation*}
and actuation matrix $\mathbf{\widetilde{G^{\mathcal{A}}}}$ is given as 
\begin{equation*}
    \resizebox{0.85\hsize}{!}{$
    \mathbf{\widetilde{G^{\mathcal{A}}}} = \begin{bmatrix} \mathbf{\widetilde{G}}[k]\\ \prod\limits_{i = k+1}^{k+1}\mathbf{\widetilde{H}}[i] \mathbf{\widetilde{G}}[k] & \mathbf{\widetilde{G}}[k+1] & &  \text{\LARGE 0} \\ \vdots & \vdots & \ddots \\ \prod\limits_{i = k+1}^{k+n-1}\mathbf{\widetilde{H}}[i] \mathbf{\widetilde{G}}[k] & \prod\limits_{i = k+2}^{k+n-1}\mathbf{\widetilde{H}}[i] \mathbf{\widetilde{G}}[k+1] & \cdots & \mathbf{\widetilde{G}}[k+n-1] \end{bmatrix}
    $}
\end{equation*}
The decision variables for this dense QP formulation are aggregated control vectors $\Vec{\mathbf{u}}$ over the $n$-step prediction horizon, $\Vec{\mathbf{u}} = \scalebox{1.3}{[} \mathbf{u}[k]^{\top} ~~ \mathbf{u}[k+1]^{\top} ~ \cdots ~ \mathbf{u}[k+n-1]^{\top} \scalebox{1.3}{]}^{\top}$. It is straightforward to define the QP objective as a sum of tracking cost and a regularization term: 
\begin{equation*}
  \resizebox{0.65\hsize}{!}{$
    \vg \Psi_{\rm qp} = \| \mathbf{\widetilde{H^{\mathcal{A}}}}  \mathbf{x}[k] + \mathbf{\widetilde{G^{\mathcal{A}}}} \Vec{\mathbf{u}} - \Vec{\mathbf{x}}^{\rm{ref}} \|_{\mathbf{Q_{\rm{qp}}}} + \| \Vec{\mathbf{u}} \|_{\mathbf{R_{\rm{qp}}}}
  $}
\end{equation*}
where $\Vec{\mathbf{x}}^{\rm{ref}} = [\mathbf{h^{\rm ref^{\top}}}, \mathbf{q_{\rm b}^{\rm ref^{\top}}}, \mathbf{q_{\rm j}^{\rm ref^{\top}}}, \mathbf{\dot{q}^{\rm ref^{\top}}_{\rm j}}]^{\top}$ are given by the desired reference, generated through offline optimization, for the entire horizon $i = k+1$ to $i = k+n$. Friction pyramid constraint, joint acceleration constraint, and normal force constraint are also added into the QP formulation to ensure physically feasible contact forces and joint accelerations. The computed contact forces and joint accelerations from the proposed MPC-QP problem are fed into the inverse dynamics (ID) controller to generate joint torques as shown in Fig.~\ref{fig:system_diagram}.

\vspace{-0.1in}
\subsection{Comparison with the original Convex-MPC}\label{sec:comparisonOriginalMPC}
Compared to the original Convex-MPC \cite{di_carlo_dynamic_2018}, our formulation added the ability to directly track centroidal momentum quantities. This offers a tighter integration with our offline trajectory generation, and fully utilizes the generated reference linear and angular momentum trajectories to faithfully execute momentum-rich motions, such as twisting jump and zero-gravity body reorientation (see Sec. \ref{sec:experiments}). Unlike the original convex-MPC controller \cite{di_carlo_dynamic_2018}, where cost heuristics must be designed to generate such motions \cite{bledt_regularized_2020}, our controller allows momentum to be designed as part of the desired reference trajectory, and directly tracked online. In addition, leveraging CMMs in the system dynamics naturally allows the inclusion of joint variables as decision variables, thus eliminating the need for any additional kinematics optimizations commonly found in such control hierarchies \cite{kim_highly_2019, wang_unified_2021}. Another benefit when compared to the original is that our formulation does not assume zero roll direction rotation of the robot. On the downside, the ability to track momentum and joint trajectories directly increases the QP problem size, and our formulation is indeed slower to compute compared to the original, for the same length prediction horizon. \Revised{Direct comparison against the original MPC is provided in Sec. \ref{sec:hardwareResults} for a $180^\circ$ twisting jump maneuver.}


The crucial assumption in our formulation is that the robot must follow the reference trajectory generated by offline planning for the approximation in Eq. (\ref{eq:LTVfullCenDyn}) to hold true. This assumption is reasonable because when the robot deviates from the offline reference, our MPC controller will generate new contact forces and joint accelerations, at a fast rate (see Sec. \ref{sec:hardwareResults}), to bring the robot state back to the reference. 

\section{Results}\label{sec:experiments}
We perform various experiments both in simulation and hardware using the alternating CEN-WBD and MPC framework. Both the Unitree A1 and MIT Mini Cheetah \cite{katz_mini_2019} were used to verify the proposed methods on quadrupedal platforms. In simulation, dynamic maneuvers such as cantering, ``parkour", and zero gravity reorientation are shown to have stable and converging solutions. In addition, hardware tests were conducted with the Mini Cheetah to verify both the offline optimization and online MPC in the real world. 
\subsection{Trajectory Generation and Simulation}

Example dynamic maneuvers generated using our offline alternating CEN-WBD optimization are presented here. \Revised{The direct transcription in CEN optimization is solved by IPOPT \cite{wachter2006implementation}, and uses a universal 0.01s timestep. For the inertia shaping example, the number of decision variables per knot point in CEN is 48, while the other examples use 42 since the inertia vector is turned off. The WBD optimization is solved by crocoddyl \cite{mastalli_crocoddyl_2020}, and applies the same amount of knot points as the CEN. The number of decision variables in WBD remains 61 (49 for states and 12 for control inputs) for all examples.} 
Aggregated solve times and residuals of both our alternating update scheme and a hierarchical one, i.e., running only one alternating iteration, are documented in Table \ref{tab:solving_time_and_residuals}. We run 3 alternating iterations for all examples. The accumulated solve times for CEN and WBD optimizations are reported separately. \Revised{We use} the residuals \Revised{to measure the consensus errors} for CoM position ($m$), EE position ($m$), and AM ($kg \cdot m^2/s$)\Revised{, which} are defined as the $l2$-norm of the error between CEN and WBD accumulated along the entire time horizon. Note that the residual for EE is also accumulated over all 4 legs. It is observed that although the solve times are longer, the residuals with more alternating iterations are significantly smaller and consensus performance is improved, which can be attributed to the benefit that one level of optimization iteratively receives feedback from the other.                           

\begin{table}[t]
    \smallskip
    \smallskip
    \caption{Solve times and residuals for trajectory generation}
    \centering
     \resizebox{\hsize}{!}{$\begin{tabular}{c c c c c}
         \hline
         Desired & Solve times 
         & Solve times 
         & Residuals for & Residuals for\\
         motions & CEN,WBD 
         & CEN,WBD 
         & CoM,EE,AM & CoM,EE,AM\\
          & (s) 
          & (Hierarchical) 
          & ($m$, $kg \cdot m^2/s$) & (Hierarchical)\\
         \hline
         Cantering & 36.8, 4.9 
         & 14.0, 3.4 
         & 1.3, 6.9, 9.9 & 1.4, 8.3, 12.7\\
         ``Parkour" & 213.0, 10.3 
         & 53.8, 7.4 
         & 5.2, 11.3, 42.9 & 53.8, 49.7, 197.2 \\
         Inertia shaping & 35.9, 48.1 
         & 4.1, 45.8 
         & 0, 12.9, 0 & 0, 290.0, 0\\
         Twisting jump & \Revised{45.3, 5.3} 
         & \Revised{2.5, 2.4} 
         & \Revised{0.7, 1.5, 6.2} & \Revised{4.15, 22.1, 47.1}\\
         \hline
    \end{tabular}
    $}
    \vspace{-0.2in}
    \label{tab:solving_time_and_residuals}
\end{table}

\begin{figure}
    \centering
    \includegraphics[width=\linewidth]{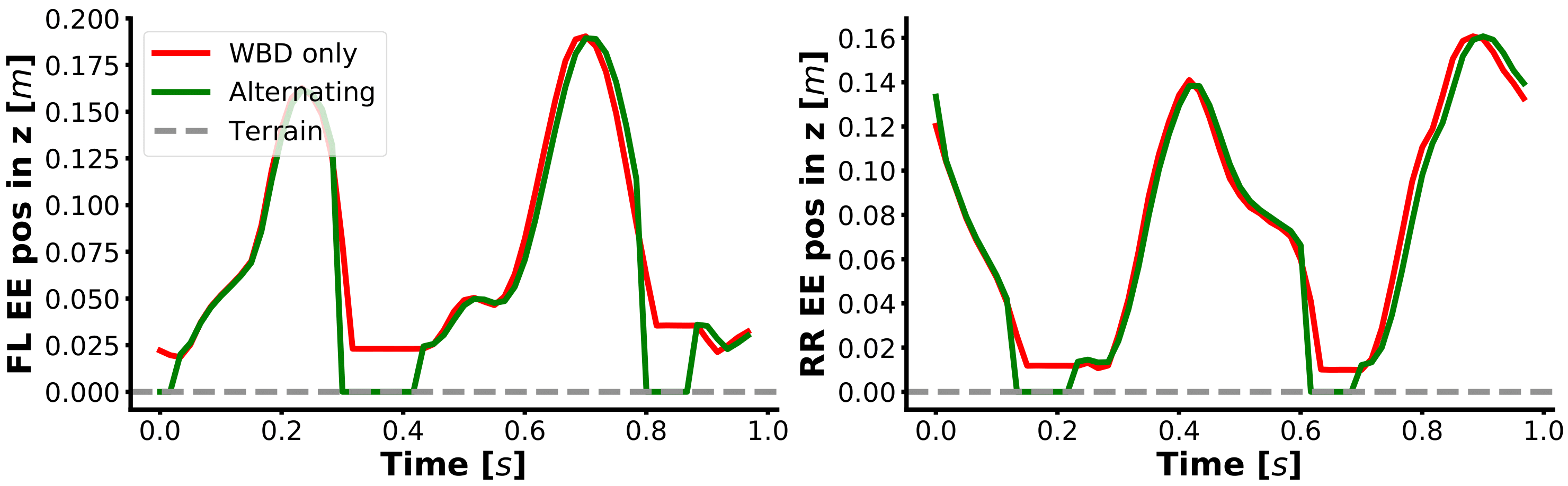}
    \caption{EE in z-axis for FL and RR legs with a flat terrain (grey), generated by using only WBD optimization (red) and our alternating one (green).}
    \label{fig:benchmark_to_wbd}
    \vspace{-0.1in}
\end{figure}

Our offline TO pipeline is able to replicate the highly dynamic cantering gait. \Revised{A one-second clip of} motion capture (mocap) data from a real animal is used to first generate the initial seed trajectory for our robot's morphology through re-targeting \cite{RoboImitationPeng20}. Then our TO pipeline uses the re-targeted inputs to produce base and joint trajectories which are not only faithful to the original mocap recording, but also physically-correct. We benchmark our solution accuracy  against that of only running WBD optimization. Fig.~\ref{fig:benchmark_to_wbd} shows a comparison of the front left (FL) and front right (FR) z-direction foot position trajectories for the cantering example. It is clear that WBD optimization by itself has difficulty constraining foot positions on the ground during stance phases, but our alternating pipeline can correctly enforce terrain constraints, which improves the overall solution accuracy . \Revised{The residuals in Table \ref{tab:solving_time_and_residuals} also reflect the improved solution accuracy, considering that most constraints to be satisfied in CEN optimization, such as the terrain constraints, are indirectly enforced in the WBD optimization through the tracking tasks.}  

A ``parkour" scenario, where the robot traverses over a large gap via inclined surfaces, using hopping gaits, is also successfully computed through our TO pipeline. We set up the CEN optimization by manually providing the desired contact sequence, timings, and a terrain map with a 2.4 s time horizon. Intermediate states, for each landing configuration on the inclined surfaces, are also specified as hard constraints to improve solver convergence rate. The cantering and ``parkour" examples are visualized in Fig.\ref{fig:sim_ex}.

\begin{figure}[htb!]
    \centering
    \includegraphics[width=\linewidth]{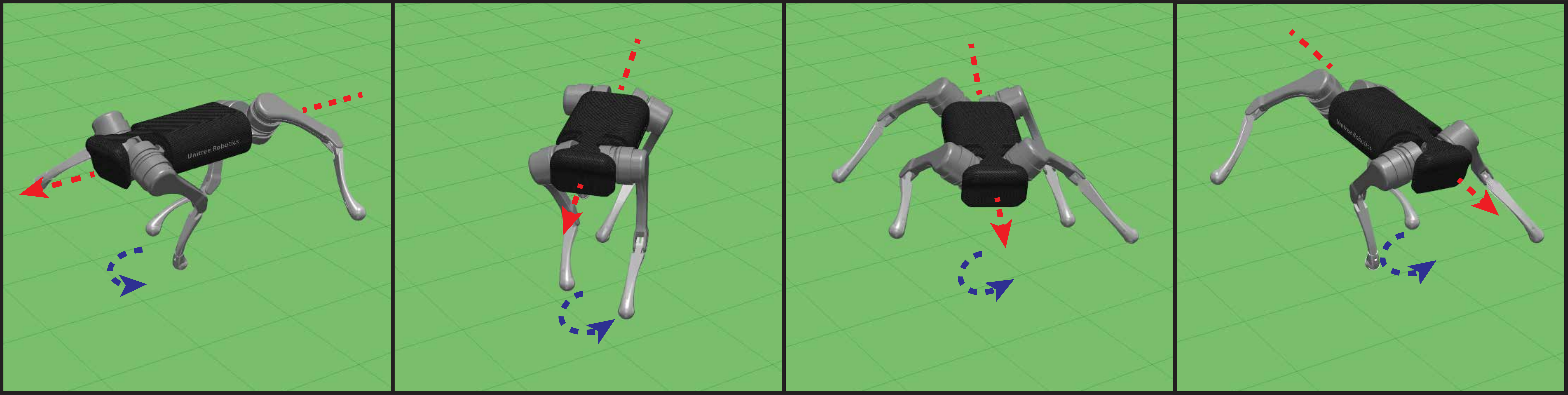}
    \caption{Simulated motion of A1 turning $90^{\circ}$ in a zero-gravity environment }
    \label{fig:zero_g_turn}
\end{figure}

To demonstrate the benefit of momentum-inertia-aware CEN optimization in the offline TO, we study the zero-gravity body reorientation problem. Fig. \ref{fig:zero_g_turn} shows the simulation result using Drake \cite{drake}. Starting from stationary initial configuration, the robot is tasked to turn its trunk $90^{\circ}$ \Revised{within two seconds}. Since no gravity nor external contacts are present in the problem setup, the robot is forced to rely on its limb motions to produce the turning behavior. In the absence of external forces, CEN optimization by itself and all other SRBM-based approaches cannot discover meaningful EE trajectories for this specific example. However, with inertia feedback from the WBD level, the two optimization levels indeed reach dynamics consensus, as shown in Table \ref{tab:solving_time_and_residuals} with the EE residual decreasing from 290 to 12.9. This further validates that inclusion of the momentum inertia parameterization at the CEN level is physically meaningful. It is also worth noting that for this example, the WBD optimization takes significantly longer than the other examples, although the total motion horizon is short. This is because during the first alternating iteration, WBD optimization lacks good tracking reference from the CEN level, due to the absence of any contact related constraints. This further illustrates that a decent tracking reference generated from CEN level helps drastically reduce the solve time for the WBD optimization.         

\begin{figure}[htb!]
    \centering
    \includegraphics[width=\linewidth]{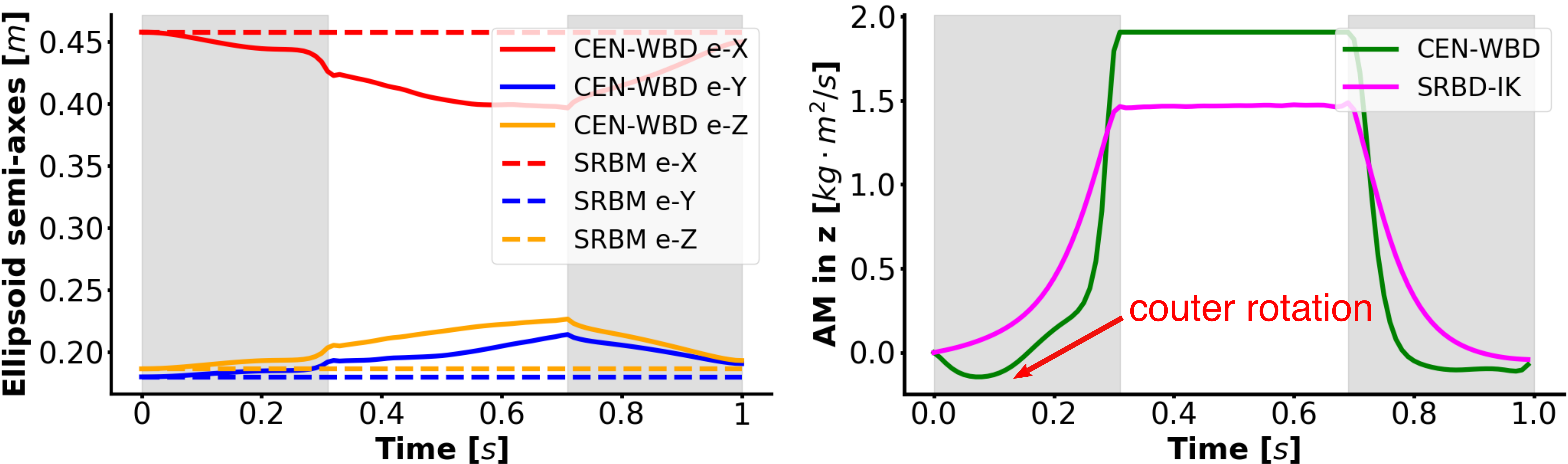}
    \caption{\Revised{Equimomental ellipsoid semi-axes changes (left) and centroidal angular momentum in z direction (right) generated by CEN-WBD and SRBM+IK for $180^{\circ}$ twisting jump.}}
    \label{fig:consensus}
\end{figure}

\begin{figure}[t!]
    \smallskip
    \smallskip
    \centering
    \includegraphics[width=\linewidth]{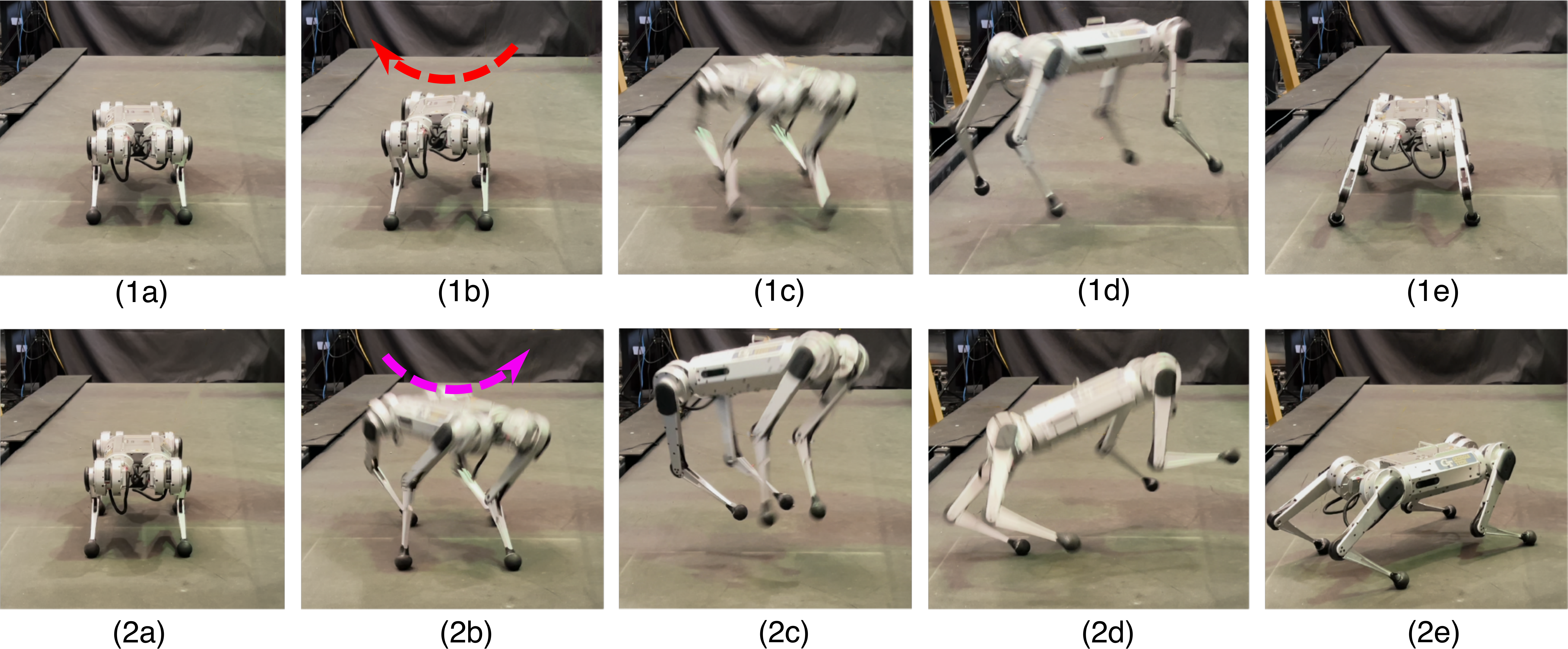}
    \caption{Demonstration of the Mini Cheetah executing a \Revised{$180^{\circ}$} twisting jump. \Revised{The first row (1a) - (1e) shows the performance on tracking the trajectory generated by our CEN-WBD, while the trajectory used in the second experiment (2a) - (2e) is from a SRBM combined with IK.}}
    \label{fig:twistingJump}
\end{figure}

The $\Revised{180}^{\circ}$ twisting jump maneuver is also studied, consisting of three contact phases: takeoff \Revised{(0-0.3s)}, aerial \Revised{(0.3-0.7s)}, and landing \Revised{(0.7-1.0s)}. The robot must gain enough AM during takeoff to achieve the goal orientation after landing. \Revised{We provide a comparison study against a SRBM with inverse kinematics (SRBM+IK)-based trajectory optimization scheme.
Different from \cite{chignoli2021online}, no hand-crafted reference trajectory is employed for SRBM+IK as a fair comparison. Centroidal AM trajectories are compared in the right sub-figure of Fig. \ref{fig:consensus}. By achieving momentum consensus through CRB inertia feedback $\v I_{\rm crb}$ from WBD, our CEN-WBD optimization leverages additional leg movements (a swift counter rotation observed in Fig. \ref{fig:twistingJump} (1b)) during the take-off phase, which leads to a larger centroidal angular momentum. The slight but non-trivial change of equimomental ellipsoid semi-axes in Fig. \ref{fig:consensus} guarantees a high-fidelity momentum trajectory to be tracked by our controller. When tested on hardware and using identical controller setups, trajectories generated using SRBM+IK fail to guide the robot to the full $180^{\circ}$ rotation, as shown in Fig. \ref{fig:twistingJump}. More hardware implementation details are described in the subsequent section.}


\subsection{Hardware Demonstration} \label{sec:hardwareResults}
We validate the tracking performance of the proposed MPC controller, by reproducing the twisting jump maneuver on the Mini-Cheetah platform. Since our MPC lacks the ability to directly optimize contact timings, a heuristics-based contact detection scheme is devised to deal with the early touchdown scenario. For each leg, touchdown is estimated by detecting an abrupt velocity change for the knee joints after the takeoff phase. For each leg, if a touchdown is early, tracking reference is immediately shifted to the point of touchdown on the offline trajectory. When all four legs land, base and momentum trajectories are also shifted to their respective points of touchdown.

\begin{figure}[htb!]
    \centering
    \includegraphics[width=\linewidth]{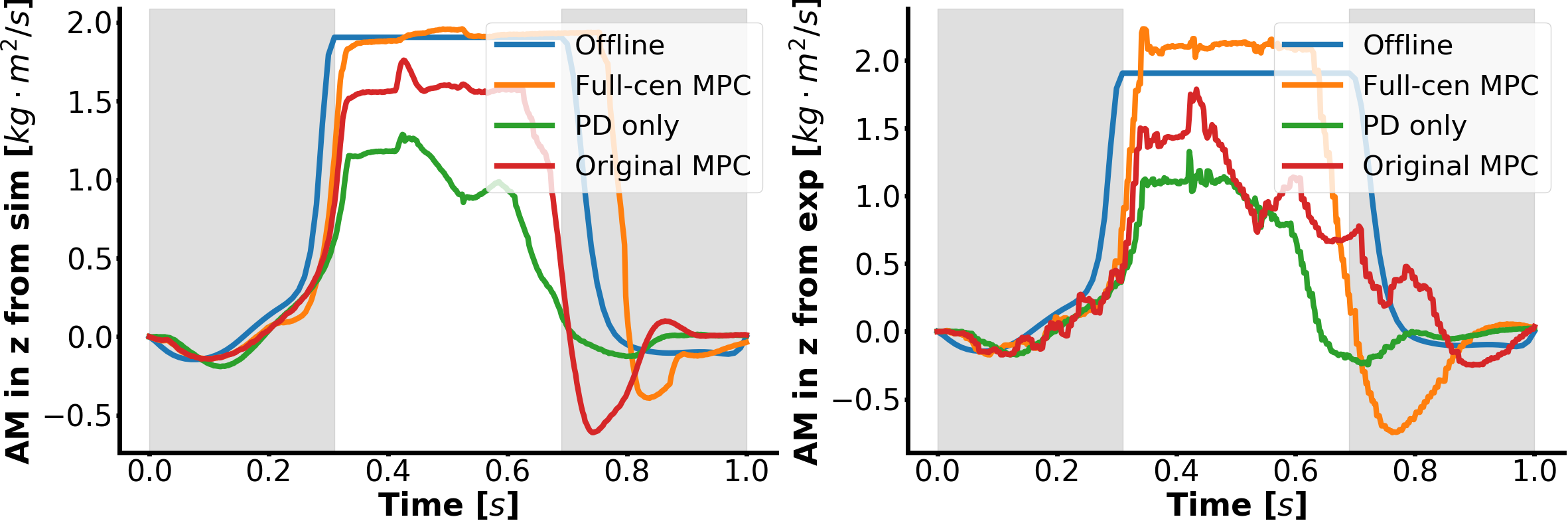}
    \caption{\Revised{Tracking performances from simulations (left) and experiments (right) for a twisting jump by comparing the offline reference trajectories and the measured data during online execution with different control strategies.}}
    \label{fig:tracking_performance_twisting_jump}
\end{figure}

Our controller is deployed directly on the Mini-Cheetah onboard computer, and updates at $100$ Hz with a prediction horizon of $50$ ms. Compare to the original convex MPC controller \cite{di_carlo_dynamic_2018}, which updates at around $30$ Hz, our version updates at a higher frequency, but with a shorter prediction horizon (up to $0.5$ s prediction horizon for the original). \Revised{Fig.~\ref{fig:tracking_performance_twisting_jump} shows the performance of both our proposed and the original convex MPC for the $180^\circ$ twisting jump maneuver. A baseline benchmark using a joint PD controller is also included in the figure. Our proposed controller gives the best tracking performance and is able to generate the largest angular momentum. In both simulation and on hardware, the original convex MPC and the PD controllers fail to achieve the full $180^\circ$ rotation, while our proposed controller successfully achieves the desired rotation. We attribute the improvement in performance over the original convex MPC to the choice of centroidal momentum as decision variables, rather than the base linear and angular velocities employed by the original. The inclusion of joint states as decision variables also allows us to better reason about joint tracking, and not relying on Cartesian PD as is the case for original convex MPC.}

\section{Conclusions and Future Work}
We presented a trajectory optimization and QP-based MPC framework for versatile and agile quadrupedal locomotion. The novelty of our proposed TO framework lied in the inclusion of equimomental ellipsoid, and an alternating scheme to achieve dynamics consensus between CEN and WBD optimizations. The MPC controller used the generated trajectory and re-computed contact forces and joint accelerations for online execution. Our proposed pipeline has been verified through a range of dynamic motions in both simulation and real world. Future works include optimizing over foot placements and contact timings in our MPC controller to realize cantering and ``parkour" on hardware.











\let\secfnt\undefined
\newfont{\secfnt}{ptmb8t at 10pt}

\bibliographystyle{IEEEtran}
\bibliography{external_ref}

\end{document}